\documentclass[10pt,journal,compsoc]{IEEEtran}
\usepackage{amsmath,amsfonts}
\usepackage{algorithmic}
\usepackage{ragged2e}
\usepackage{array}
\usepackage{textcomp}
\usepackage{stfloats}
\usepackage{url}
\usepackage{verbatim}
\usepackage{graphicx}
\usepackage{booktabs}
\usepackage{float}
\usepackage{subcaption}
\usepackage{cite}
\usepackage[colorlinks]{hyperref}
\usepackage[ruled,linesnumbered]{algorithm2e}
\usepackage{animate}
\usepackage{multirow}
\usepackage{nth}
 
\usepackage{multicol}

\def\BibTeX{{\rm B\kern-.05em{\sc i\kern-.025em b}\kern-.08em
    T\kern-.1667em\lower.7ex\hbox{E}\kern-.125emX}}
    
\usepackage{balance}

\begin{document}
\title{AnchorCrafter: Animate Cyber-Anchors Selling Your Products via Human-Object Interacting Video Generation  
}
\author{Ziyi Xu,~Ziyao Huang,~Juan Cao,~Yong Zhang,~Xiaodong Cun,~Qing Shuai,~Yuchen Wang,\\~Linchao Bo,~Fan Tang
\IEEEcompsocitemizethanks{
\IEEEcompsocthanksitem Z. Xu, Z. Huang, J. Cao, Y. Wang and F. Tang are with the Institute of Computing Technique, Chinese Academy of Sciences, Beijing, China. 
\IEEEcompsocthanksitem Y. Zhang is with Meituan, Beijing, China.
\IEEEcompsocthanksitem X. Cun is with Great Bay University, Guangdong China. 
\IEEEcompsocthanksitem Q. Shuai and L. Bao are with Tencent AI Lab. 
}

}


\IEEEtitleabstractindextext{
\begin{abstract}
\justifying
The generation of anchor-style product promotion videos presents promising opportunities in e-commerce, advertising, and consumer engagement. 
Despite advancements in pose-guided human video generation, creating product promotion videos remains challenging. 
In addressing this challenge, we identify the integration of human-object interactions (HOI) into pose-guided human video generation as a core issue. 
To this end, we introduce AnchorCrafter, a novel diffusion-based system designed to generate 2D videos featuring a target human and a customized object, achieving high visual fidelity and controllable interactions. 
Specifically, we propose two key innovations: the HOI-appearance perception, which enhances object appearance recognition from arbitrary multi-view perspectives and disentangles object and human appearance, and the HOI-motion injection, which enables complex human-object interactions by overcoming challenges in object trajectory conditioning and inter-occlusion management. 
Extensive experiments show that our system improves object appearance preservation by 7.5\% and doubles the object localization accuracy compared to existing state-of-the-art approaches. It also outperforms existing approaches in maintaining human motion consistency and high-quality video generation. 
Project page including data, code, and Huggingface demo: \url{https://github.com/cangcz/AnchorCrafter}.

\end{abstract}
}

\maketitle

\section{Introduction}

\label{sec:intro}
\IEEEPARstart{V}{iewing} unboxing videos and live-streamed product promotions by content creators and live streamers\textendash{collectively referred to as anchors}\textendash{}on platforms such as YouTube, TikTok, Douyin, etc, has become an integral part of the online shopping experience. Recent advancements in computer graphics, including video generation~\cite{24makeyourvideo,25MotionCrafter} and human animation~\cite{qu2025controllable,shao2024human4dit}, have made it possible to automate the creation of such unboxing and promotional content. However, achieving high-fidelity temporal consistency, object realism, and controllable motion generation remains a formidable challenge.

\begin{figure}[t]
\newcommand{\galleryfigurewidth}{0.22}
\centering
    \begin{minipage}[t]{\linewidth}
    \centering
        \begin{minipage}{\galleryfigurewidth\linewidth}
            \centering
            \includegraphics[width=\linewidth]{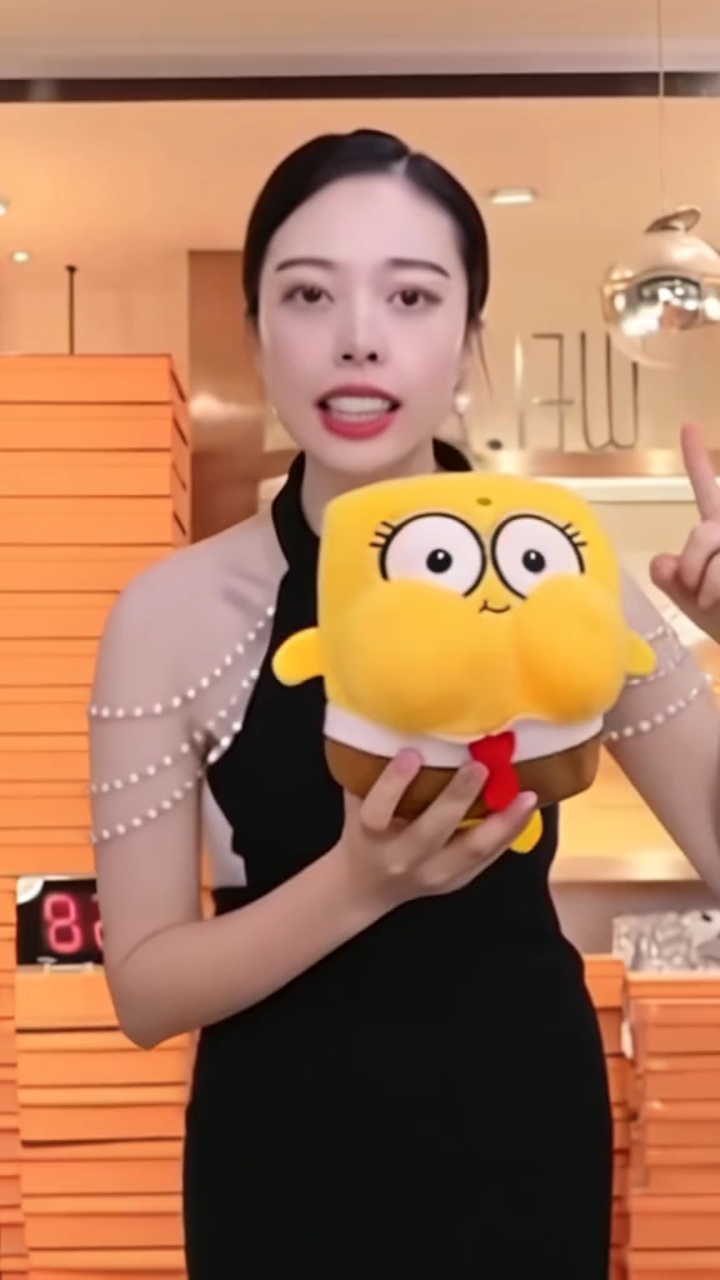}
            \subcaption*{\scriptsize Reference Input}
        \end{minipage}
        \begin{minipage}{\galleryfigurewidth\linewidth}
            \centering
            \includegraphics[width=\linewidth]{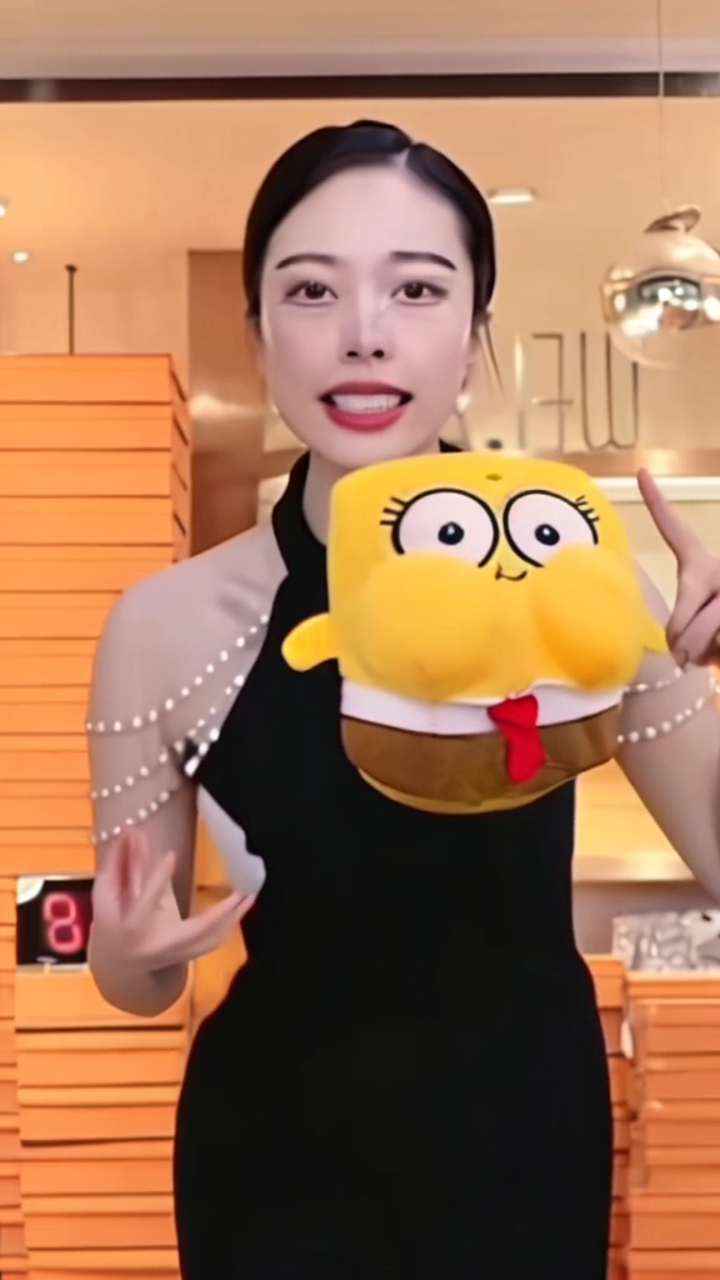}
            \subcaption*{\scriptsize  MimicMotion}
        \end{minipage}
        \begin{minipage}{\galleryfigurewidth\linewidth}
            \centering
            \includegraphics[width=\linewidth]{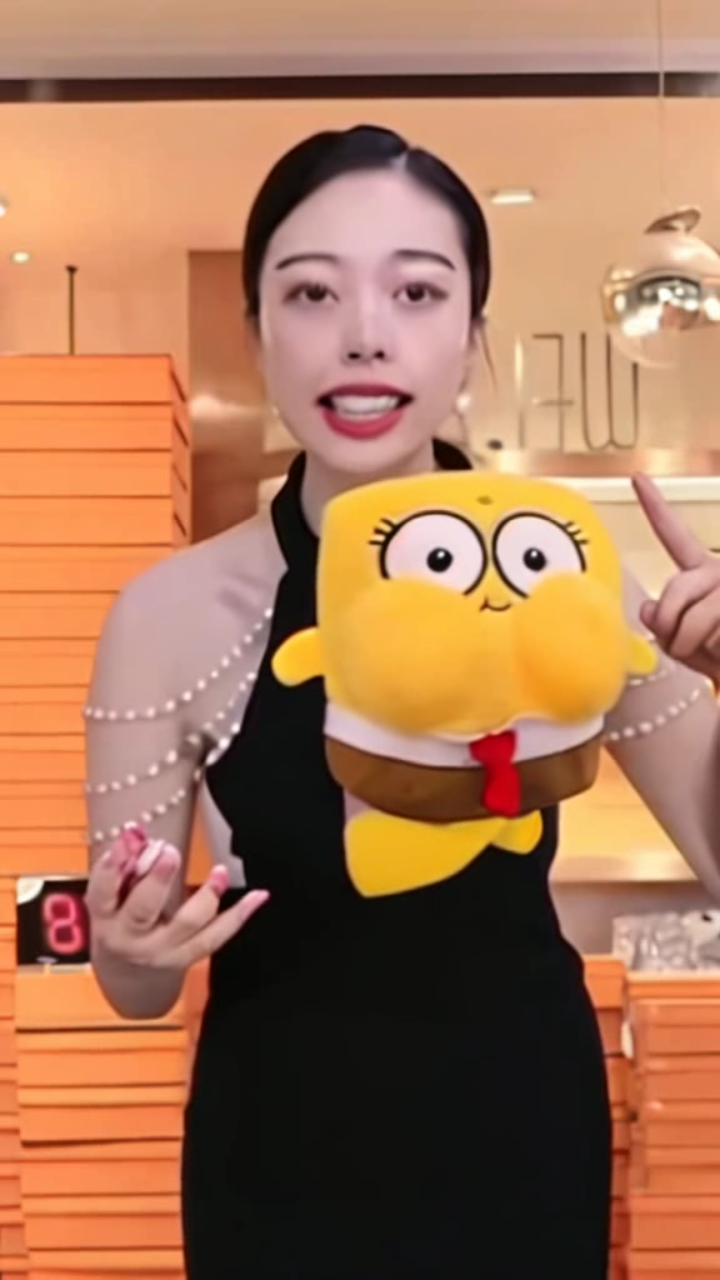}
            \subcaption*{\scriptsize StableAnimator}
        \end{minipage}
        \begin{minipage}{\galleryfigurewidth\linewidth}
            \centering
            \includegraphics[width=\linewidth]{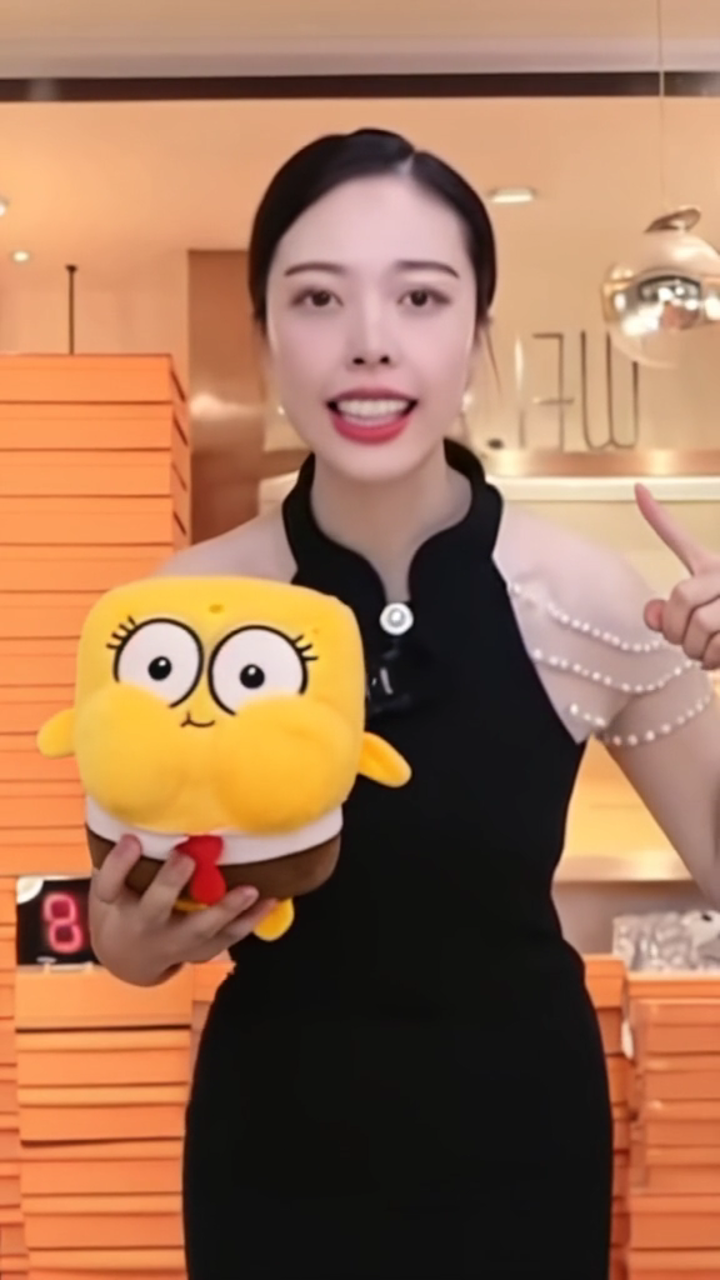}
            \subcaption*{Ours}
        \end{minipage}
    \end{minipage}

\caption{
Existing methods accurately follow human poses but struggle with realistic hand-object interactions, often misinterpreting the object as part of the human, leading to static animations. In contrast, our approach ensures natural and dynamic movement by precisely synthesizing human-object interactions while preserving object appearance.
}
\label{fig:1_Intro}
\end{figure}

Pose-guided human video generation~\cite{hu2024animateanyone, xu2024magicanimate, zhang2024mimicmotion, huang2024makeyouranchor,tu2025stableanimator, chang2025x-dyna} aligns closely with anchor-style product promotion videos. 
Existing diffusion-based methods generate temporally consistent, high-fidelity human videos using pose and appearance references. 
Make-Your-Anchor~\cite{huang2024makeyouranchor} pioneered personalized anchor-style video generation, while methods like MimicMotion~\cite{zhang2024mimicmotion} and StableAnimator~\cite{tu2025stableanimator} animate static human images with precise poses, adaptable for anchor-style video generation. 
However, due to the lack of effective human-object interaction (HOI) capabilities~\cite{xue2024hoi, chen2024virtualmodel, fan2025rehold}, these methods are unable to generate human-driven product demonstrations. As illustrated in Fig.~\ref{fig:1_Intro}, objects are often treated as static textures, which restricts interactive realism and hinders the modeling of complex hand-object interactions.


Recent studies on HOI generation~\cite{xue2024hoi, chen2024virtualmodel,fan2025rehold} primarily focus on hand-centric videos or image domains, which do not offer the degrees of freedom necessary for anchor-style product promotion videos. Therefore, seamlessly integrating HOI into pose-guided human video generation is essential for achieving natural and dynamic human-object interactions in synthesized videos.

\begin{figure*}
    \captionsetup{type=figure}
    \includegraphics[width=\linewidth]{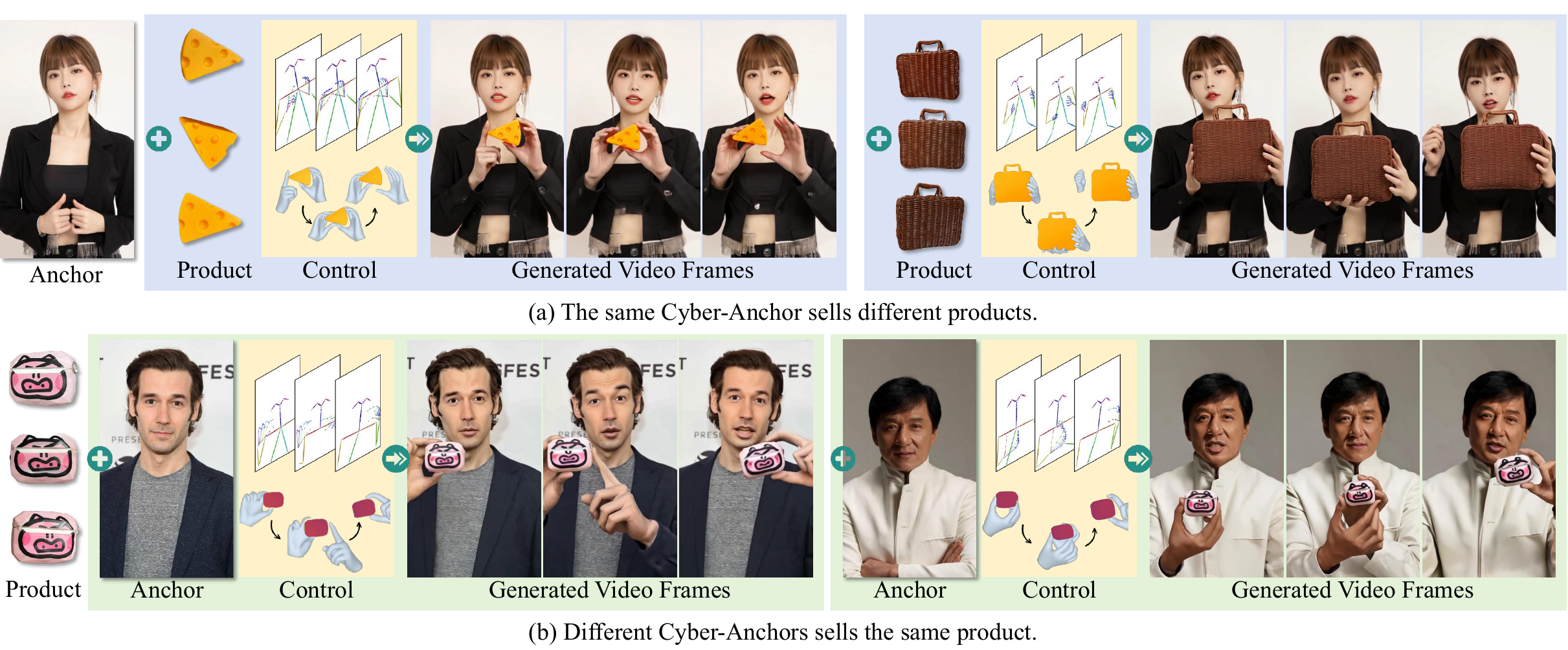}
    \vspace{-5mm}
    \caption{We propose AnchorCrafter, a diffusion-based human video generation framework for creating high-fidelity anchor-style product promotion videos by animating reference human images with specific products and motion controls. By incorporating human-object interaction into the generation process, AnchorCrafter achieves high preservation of object appearance and enhanced interaction awareness.}
   \label{fig:teaser}
\end{figure*}

To generate anchor-style product promotion videos, we introduce AnchorCrafter, a novel object-centric customization framework. 
Unlike person-centric approaches such as Make-Your-Anchor~\cite{huang2024makeyouranchor}, our method tailors a video diffusion model by collecting a one-minute interaction video of a specific object, enabling arbitrary anchors to naturally interact with that object in diverse poses.
Specifically, we refine human and object representations via HOI-appearance perception, which integrates multi-view features and adopts a decoupled architecture to effectively disentangle human and object appearances.
However, without precise motion control, objects remain disconnected from human interactions. To achieve natural and coordinated object behavior, we propose the HOI-motion injection module, which leverages depth maps and 3D hand meshes to provide fine-grained motion guidance to the model.  In addition, interaction artifacts are alleviated through the adoption of occlusion-handling strategies.
Furthermore, conventional training objectives often underrepresent the subtle dynamics of hand-object interactions, which undermines the realism of generated content. To enhance interaction fidelity, we introduce an HOI-region reweighting loss that strategically emphasizes interaction-critical regions during training, leading to reinforced fine-grained object representations and improved realism in modeled interactions.
We conduct extensive experiments on real-world objects and demonstrate that AnchorCrafter achieves object appearance preservation and interaction awareness results, which current approaches do not accomplish. Moreover, we collected and open-sourced a digital human interaction dataset to fill the gap in the field of interactive digital human generation.
\noindent In summary, our contributions are as follows:
\begin{itemize}
\item We propose AnchorCrafter, a novel system incorporating human-object interaction into pose-guided human video models and generating realistic anchor-style product promotion videos by animating reference anchors. 
\item We propose an HOI-appearance perception that utilizes a novel multi-view feature fusion and decoupled injection structure to achieve object appearance preservation, an HOI-motion injection to condition object motion and handle inter-occlusion, and an HOI-region reweighting loss to enhance object details.
\item We conducted quantitative and qualitative evaluations to demonstrate the effectiveness of AnchorCrafter, comparing it with state-of-the-art diffusion-based human video generation and editing methods.
\end{itemize}


\section{Related Work}
\subsection{Pose-Guided Human Video Generation}
Pose-guided human video generation utilizes sequential poses to control human motion in generated videos, commonly applied in scenarios such as dance and speech video generation.
With rapid advancements in diffusion models~\cite{ho2020ddpm, rombach2022ldm} and pose-controlled human image generation~\cite{li2020pona, zhang2021pise}, pose-guided human video generation has achieved remarkable progress.

Common practices leverage ControlNet~\cite{zhang2023controlnet}, a parallel control branch attached to diffusion UNet, to inject human skeletal motion sequences~\cite{wang2024disco,xu2024magicanimate,tu2024motioneditor, huang2024makeyouranchor, qu2025controllable}, thereby achieving advanced pose guidance.
AnimateAnyone~\cite{hu2024animateanyone} proposed ReferenceNet to control human appearance, which injects information through the self-attention mechanism.
Champ~\cite{zhu2024champ} enhances motion control by combining four different types of poses. MimicMotion~\cite{zhang2024mimicmotion} engineered a lightweight pose guidance network to incorporate human pose conditions, and employed confidence-aware pose guidance to enhance the quality of hand motion synthesis. To further improve facial detail embedding, StableAnimator~\cite{tu2025stableanimator} introduces a Face Encoder, enhancing precision in facial synthesis.

Recent studies have explored replacing UNet with diffusion transformer(DiT) to achieve better temporal consistency in video generation. Based on the DiT framework, HumanDiT~\cite{gan2025humandit} employs prefix latent reference strategies, ensuring visual consistency across extended sequences. UniAnimate-DiT~\cite{wang2025unianimate}, built upon Wan2.1~\cite{vace}, incorporates Low-Rank Adaptation~\cite{hu2022lora} to facilitate efficient digital human synthesis while reducing computational overhead. Similarly to UNet-based approaches, both models leverage pose sequences to control human motion within generated images. However, despite these advancements, DiT-based frameworks still require large-scale datasets for training to uphold video quality and coherence, posing challenges in practical deployment.

These studies demonstrate that by incorporating human poses such as OpenPose~\cite{cao2017openpose}, DensePose~\cite{guler2018densepose}, {SMPL-X}~\cite{pavlakos2019smplx}, control over human movement can be achieved. However, the important human-object interaction generation is often overlooked, despite its vital importance for real-world applications such as anchor-style product promotion video generation. In contrast, AnchorCrafter incorporates HOI into human video generation to address this issue.

\subsection{Human-Object Interaction Generation}
The goal of HOI is to generate visual content that accurately represents the relationships between humans and objects. In the domain of 3D reconstruction, EasyHOI~\cite{liu2025easyhoi} enables hand-object interaction reconstruction from a single image, whereas DICE~\cite{25DICE} facilitates single-image reconstruction of hand-face interactions. In the field of motion generation~\cite{Wu2024Doodle,ji2025sport}, some studies focus on interactive motion synthesis, aiming to generate human-object interactions with enhanced realism.
InterDiff~\cite{xu2023interdiff}, TeSMo~\cite{yi2024TeSMo} and HOIAnimator~\cite{Song2024HOIAnimatorGT} predict human motions during interactions with objects based on 3D object representations. PMP~\cite{bae2023pmp} learns to use part-wise motion priors to physically interact with environments. IMoS~\cite{ghosh2023imos} generates full-body human and 3D object motions from textual input, but its focus is primarily on hand-grasping small objects. TextIM synthesizes human interactive motions with part-level semantic accuracy by aligning textual descriptions with interactive body movements.

However, human-object interaction remains relatively underexplored in the field of content generation. HOI-Swap~\cite{xue2024hoi} zeroes in hand-centric videos, utilizing a two-stage editing technique to swap the object in hand while keeping the hand-object interactions. ReHold~\cite{fan2025rehold}proposes the HOI Restoration Module, which injects hand and object details into the model to enhance hand-object interaction. VirtualModel~\cite{chen2024virtualmodel} stands as the most relevant work in literature, albeit within the domain of image generation. It excels in synthesizing interactive images of characters by leveraging input objects and human poses.
Existing methods primarily focus on HOI image generation or object swapping in hand-centric videos, leading to the absence of the ability to generate videos with sufficient degrees of freedom, particularly videos that include the entire human body. 
In contrast, our approach integrates HOI capabilities into a human video generation model, enabling the generation of high-quality, controllable anchor-style product promotion videos.

\section{System Setting}
\label{sec:method.overview}

\begin{figure}
  \centering
   \includegraphics[width=1.0\linewidth]{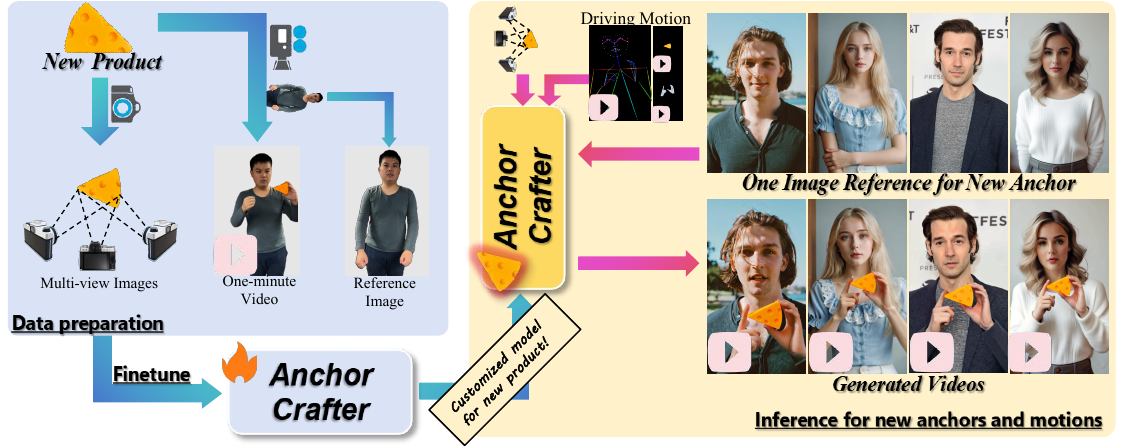}
   \caption{Fine-tuning for new products and inference for new anchors. Fine-tuning the model with a one-minute video to achieve a customized model for new objects. After fine-tuning, our method could generate arbitrary anchor videos selling the product with various unseen motions. }
\label{fig:finetune}
\end{figure}

\noindent\textbf{Overview.}
Driven by cutting-edge progress in pose estimation~\cite{yang2023dwpose,liu2025motions} and enhanced by access to large-scale human datasets~\cite{jafarian2021tiktok, wang2024humanvid,li2025mvhumannet++,pang2024dreamdanceanimatinghumanimages}, techniques~\cite{zhang2024mimicmotion,wang2024disco,xu2024magicanimate,wang2024humanvid} for producing high-fidelity, controllable human videos have become increasingly mature.
However, the absence of open-source human-object interaction datasets poses a significant challenge to generating anchor-style product promotion videos requiring accurate reconstruction of objects. 
To better address customized product interactions, we frame the task as a learning paradigm for customized products: learns fine-grained texture and appearance from a one-minute interaction video with the product. By leveraging this learned representation, the model generalizes effectively, producing interaction videos with diverse human appearances and precisely controlled motions.
The learning framework of our system consists of two stages: training on diverse datasets to capture the distribution of HOI videos, and fine-tuning on specific objects, enabling more precise and adaptable interaction modeling.

\noindent\textbf{Training.} 
During training, each source video $X=\{x^{1:f}\}$ with $f$ frames enables the model to learn the target object $O$ and its interaction patterns with humans. To ensure accurate appearance modeling, we use a reference image $I_H$ for the anchor and multi-view images $I_O = \{i_O^1, i_O^2, i_O^3\}$ for the object.
We employ multiply comprehensive conditions to control the HOI motions, including the sequences of human skeletons $P = \{p^{1:f}\}$, 3D hand meshes $H = \{h^{1:f}\}$ and depth maps of the object $D = \{d^{1:f}\}$. 
The system then generates a sequence of frames $Y = \{y^{1:f}\}$ featuring a target human and customized object with controlled interactions. To improve interaction modeling, we collect a foundational dataset of human-object interactions. Given the diversity of objects and dataset constraints, our primary focus is refining interaction control rather than capturing intricate texture details.

\noindent\textbf{Fine-tuning.} During fine-tuning, we collect videos exceeding one minute in length for each object requiring demonstration, as illustrated in Fig.~\ref{fig:finetune}. 
Benefiting from the training phase, where the model learns human-object interaction concepts, the fine-tuning stage requires only a short video to grasp object textures. This enables the model to generate previously unseen poses and novel anchor interactions. We maintain that collecting one-minute videos is both practical and contributes to a more realistic representation of the object.

\noindent\textbf{Inference.} In the inference stage, we use HOI motions performed by actors to drive the generation of anchor videos.
Actors demonstrate the required interaction actions by holding the products $O$ learned during fine-tuning, and the extracted control conditions $P$, $H$, and $D$ will serve as inputs to the system. Subsequently, our model can generate product-promotion videos for any unseen anchor image $I_H$.

\section{Methodology}
\label{sec:method}
\begin{figure}
  \centering
   \includegraphics[width=1.0\linewidth]{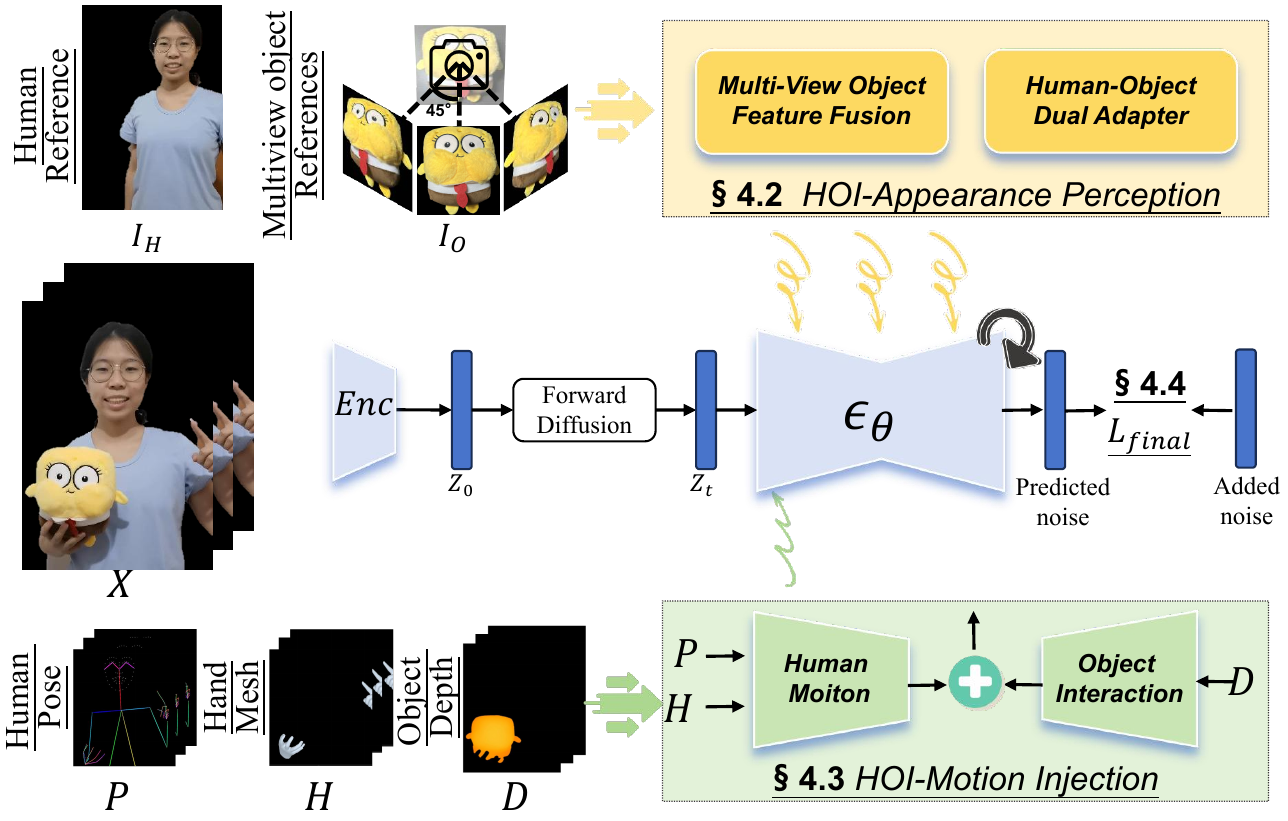}
   \caption{
Training pipeline for AnchorCrafter: Based on a video diffusion model, AnchorCrafter injects human and multi-view object references into the video via HOI-appearance perception. The motion is controlled through HOI-motion injection, with the training objective reweighted in the HOI region.
   }
  \label{fig:pipeline}
\end{figure}

As illustrated in Fig.~\ref{fig:pipeline}, our system's pipeline consists of two core components: HOI-appearance perception and HOI-motion injection. We first introduce the video diffusion model architecture utilized in AnchorCrafter in Sec.~\ref{sec:method.vdm}.
To enhance the reconstruction of both human and object appearances, we design HOI-appearance perception (Sec.~\ref{sec:method.HOAM}), which employs an appearance-decoupling strategy to ensure accurate representation. To regulate the movement of humans and objects, we integrate HOI-motion injection (Sec.~\ref{sec:method.IIM}), which introduces precise control signals into the system. Additionally, to refine interaction fidelity, we introduce HOI-region reweighting loss (Sec.~\ref{sec:method.strategies}), which strategically enhances hand-object interaction details during inference.
\subsection{Video Diffusion Model}
\label{sec:method.vdm}
AnchorCrafter is based on a video diffusion model~\cite{blattmann2023svd} architecture, comprising a diffusion UNet~\cite{ronneberger2015unet, rombach2022ldm} with temporal layers, denoted as $\epsilon_{\theta}$, and a variational autoencoder (VAE)~\cite{kingma2013vae} composed of an encoder $Enc$ and a decoder $Dec$ for compressing and uncompressing the video frames. During training, the video sequence $X$ is encoded into the latent space as $Z_0=Enc(X)=\{z_0^{1:f}\}$.
The core training objective for the video diffusion model is:
\begin{equation}
    L_{diff}=\mathbb{E}_{\epsilon\sim N(0,I),Z,c,t}[||\epsilon-\epsilon_{\theta}(Z_t,c,t) ||^2_2 ],
    \label{equation:diffusionloss}
\end{equation}
where $c$ is conditional signal, $Z_t = \{z^{1:f}_t\}$ is the latent feature diffused from $Z_0$ through a deterministic Gaussian process over $t$ timesteps by adding noise $\epsilon$.
During inference, the initial noises $U=\{u^{1:f}\}$ sampled from Gaussian noise are denoised with $T$ timesteps to get the estimated $\hat{Z_0}$, and the output video with $Y=Dec(\hat{Z_0})=\{y^{1:f}\}$.

\subsection{HOI-Appearance Perception}
\label{sec:method.HOAM}
\begin{figure}
  \centering
   \includegraphics[width=1.0\linewidth]{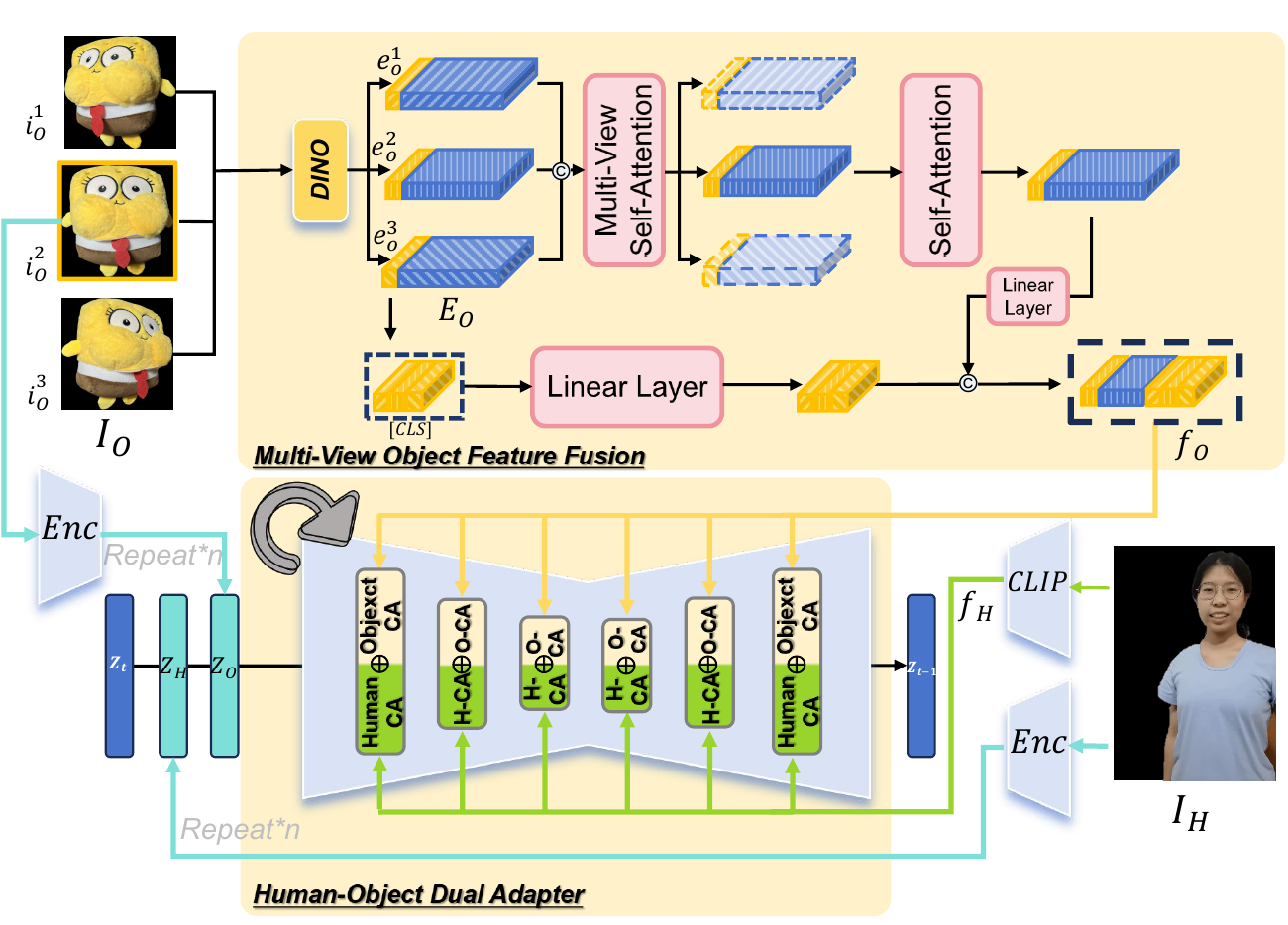}
   \caption{
   HOI-appearance perception: The feature of the target object $f_O$ is extracted through multi-view object feature fusion and combined with the human reference feature $f_H$ within a human-object dual adapter to achieve improved disentanglement results.
   }
  \label{fig:method}
\end{figure}

As illustrated in Fig.~\ref{fig:method}, the HOI-appearance perception design focuses on extracting appearance features for humans and objects, which are then integrated into the backbone network. Referring to previous approaches~\cite{hu2024animateanyone, wang2024disco, zhang2024mimicmotion, xing2025dynamicrafter}, we integrate global context and local details from input images for human appearance.


Specifically, the VAE encoder $Enc$ maps human images to a latent space compatible with the diffusion model $Z_H = Enc(I_H)$, which preserves the local detail. Subsequently, $Z_H$ is repeated $f$ times and concatenated with the UNet input $Z_t$. Meanwhile, the CLIP~\cite{radford2021clip} features of the human image are extracted as $f_H = CLIP(I_H)$ for global representation.

To accurately depict objects, especially the promoting product, the model must perceive their shape and texture from multiple perspectives. 
Instead of a single reference image, we utilize multi-view object reference images $I_O=\{i_O^1, i_O^2, i_O^3\}$ captured from 45° left, front, and 45° right angles.
Similar to human reference, the VAE feature of the front reference $i_O^2$ is extracted as $Z_O = Enc(i_O^2)$ and then repeated $f$ times and concatenated with $Z_t$.
Different from the process for human reference, we propose multi-view object feature fusion to extract richer information from multi-view images of the input object images. 

\subsubsection{Multi-View Object Feature Fusion}

We introduce the multi-view object feature fusion to understand object appearance inherently.
The process starts by feeding the multi-view object reference $I_O$ into the pre-trained DINOv2-large model~\cite{oquab2023dinov2}, resulting in the extraction of embedding $E_O=\{e_O^1, e_O^2, e_O^3\}\in R^{n \times m}, n=1370, m=1024$. Then, $E_O$ is processed through two distinct branches, as illustrated in Fig.~\ref{fig:method}. In the first branch, we designed a multi-view self-attention layer to merge object features from three perspectives, where three object embeddings of shape \( n \times m \) from DINO are concatenated along the $n$ channel and then a self-attention is performed. Subsequently, we extract the \( e_O^2 \) of the embedding, perform self-attention, and employ a linear network to project the features into a reduced dimensional space of \(16\times m\). In the other branch, we concatenate the first [CLS] feature from $E_O$ as a vector of size $R^{3 \times m}$ for global representation, then pass through a linear layer.
Finally, concatenate two branches' outputs as the final multi-view object representation, $f_O \in R^{19 \times m}$. This module extracts object features precisely from images captured from three perspectives, extracts 3D consistency features from multi-view images~\cite{tang2025mvdiffusion++}, and facilitates the final reproduction of objects.

\subsubsection{Human-Object Dual Adapter}
We observed that directly integrating object features $f_O$ into UNet along with human features $f_H$ leads to the appearance entanglement of the objects and humans.
We propose a human-object dual adapter by replacing the cross-attention layers in each block of the diffusion UNet to achieve better disentanglement between humans and objects.
The human feature $f_H$ extracted from CLIP~\cite{radford2021clip} and the object feature $f_O$ obtained by multi-view object feature fusion are each fed into a cross-attention layer, and the formula can be written as:
\begin{align}
    Human CA& := Softmax(\frac{Q K_{H}^T}{\sqrt{d}})\cdot V_H,
\end{align}
and
\begin{align}
    Object CA& := Softmax(\frac{Q K_{O}^T}{\sqrt{d}})\cdot V_O,
\end{align}
where $Q = W_Q \cdot Z_t$, $K_H =W_{KH} \cdot f_H$, $K_O =W_{KO} \cdot f_O$, $V_H =W_{VH} \cdot f_H$, $V_O =W_{VO} \cdot f_o$, $Z_t=\{z^{1:f}_t\}$ is the input latent frames on diffusion timestep $t$ and $W_Q$, $W_{KH}$, $W_{KO}$, $W_V$ are learnable weights for attention modules. Obtain the final feature set by summing the output features from the  $Human CA$ and $Object CA$ layers. Dual adapter effectively extracts features of objects and humans while achieving the disentanglement of the two entities.
\subsection{HOI-Motion Injection}
\label{sec:method.IIM}
We propose an HOI-motion injection module that explicitly models hands and occlusion to address the challenge of generating specified interactive motions.

A critical challenge in conditioning the movement of objects is providing an unambiguous representation of their orientation and positional trajectory in 3D space. To address this, the depth map $D$ serves as the primary input for the object's trajectory. Estimate and crop the depth map of the target object, then process it through a lightweight nine-layer convolutional network. Subsequently, the resulting features are element-wise added to the output of the first convolutional layer of the UNet. Due to the suboptimal performance of existing pose extractors for complex hand movements, particularly under occlusion, this study further considers the occlusions occurring between hands and objects during interactions. In addition to the commonly used human skeleton $P$ for controlling the overall human pose, 
we extract the 3D mesh sequences $H$ of the human hands and mask the corresponding 3D mesh where the object occludes the hands. The two conditions are concatenated and processed through a convolution network with the same structure as the object interaction but with unshared parameters. The features extracted by this network are subsequently added to the features within the UNet.

We observe that spatial differences between the input pose sequence $P$, $H$, and $D$ and the reference human image $I_H$ affect the generated results. 
For instance, when reference $I_H$ depicts a person closer to the camera, while the input pose $P$ is farther from the camera and includes more body parts, the input pose does not correspond to the reference image, leading to distortion in the consistency and accuracy of the generated video. To address this problem, we estimate a similarity matrix from the first frame of $P$ to the pose of $I_H$, and then the similarity matrix is performed on all motion conditions including $P$, $H$, and $D$ to match the spatial position with $I_H$.

\subsection{HOI-Region Reweighting Loss}
\label{sec:method.strategies}
By constructing modules for HOI, our model enables control over the generated humans, objects, and their interactions. Accordingly, based on the diffusion model's loss function described in Eq.~\ref{equation:diffusionloss}, our training objective can be formulated with $c = \{P,H,D,I_H,I_O\}$.
However, we observed that standard training loss caused the model to fail to learn object appearances adequately. 

To mitigate this, we propose a HOI-region reweighting loss, which enhances the model's attention to interaction regions throughout the training process:
\begin{equation}
    L_{object} = \eta \frac{S}{S_{obj}+S_{hand}} M_{inter} \odot L_{diff},
\end{equation}
where $S$, $S_{obj}$, $S_{hand}$ is the area of the whole image, object, and hands, $M_{inter}$ indicates the mask of the interaction region consisting of object and hands, and $\eta$ is a hyperparameter.
Using the inverse of the image area occupied by the interaction region as a weight, the model can assign higher training importance to the interaction area.
The final loss is:
\begin{equation}
    L_{final} = (1-M_{inter}) \odot L_{diff} + L_{object}.
\end{equation}
After applying the final loss, pose-driven capability across diverse human appearances is preserved, and objects are learned precisely.

\section{Experiments}
\label{sec:expr}
This chapter provides a comprehensive evaluation of our system. We first introduce the dataset and experimental setup (Sec.~\ref{sec:Settings}), followed by benchmarking against STOAs to assess appearance preservation, motion control, and interaction realism (Sec.~\ref{sec:Main Results}). 
To further validate our design choices, we conduct ablation studies on HOI-appearance perception, HOI-motion injection, loss reweighting, and fine-tuning strategies (Sec.~\ref{sec:Ablation Study}), along with a user study (Sec.~\ref{sec:User Study}). Finally, we discuss the scalability of our method, which can handle non-rigid and similarly shaped objects (Sec.~\ref{sec:Discussions}).
\subsection{Experimental Settings}
\label{sec:Settings}
\subsubsection{Dataset}


For training, due to the lack of an open-source dataset specifically suited to our task, we constructed a custom dataset comprising two types of videos: human-only videos and human-object interaction videos. 

\begin{itemize}
\item \textbf{Human-only videos.} The human-only subsets comprising a total of 9,906 videos originate from two open-source datasets: HumanVid~\cite{wang2024humanvid} and DreamDance~\cite{pang2024dreamdanceanimatinghumanimages}. HumanVid includes a variety of resolutions and encompasses both internet-sourced and generated data. We filtered the HumanVid to retain only vertical videos from the internet-sourced subset featuring a single person, resulting in a collection of 5,848 videos. Meanwhile, DreamDance is a human dance dataset providing URLs for each video. Due to access restrictions, we successfully downloaded 4,058 videos.

\item \textbf{Human-object interaction videos.} The human-object interaction subset includes 356 real-world videos recorded using mobile devices, as well as 44 livestream recordings collected from online platforms. These videos feature 11 participants interacting with 286 distinct objects, primarily comprising common household items such as toys, cups, and boxes. Each video has a duration of approximately 20 seconds at 30 FPS, capturing natural and spontaneous human-object interactions. To ensure diverse and complete object representation, we photograph objects from three perspectives: frontal, 45° left rotation, and 45° right rotation. For a small number of objects lacking three-view coverage, duplicated images were used to approximate multi-view input.

\end{itemize}


For testing, we compiled a dataset of 80 videos, featuring interactions between eight individuals sourced from internet and five distinct objects. Each object includes approximately one minute of interaction footage for fine-tuning, along with two eight-second clips for evaluation. The individuals engage with the objects in diverse poses, ensuring a broad range of interaction scenarios. The training and testing datasets have been made public.

\noindent\textbf{Pre-processing.} For each video, we use DWPose~\cite{yang2023dwpose} to extract human skeletal pose sequences, following~\cite{zhang2024mimicmotion}. For the videos we recorded, we additionally employ HaMeR~\cite{pavlakos2024hamer} to extract hand 3D mesh sequences, SAM2~\cite{ravi2024sam2} to extract object motion masks, and ViTA~\cite{xian2023vita} to produce depth maps. For the human-only videos, since they do not involve objects and rapid hand movements result in blurring, we use black pixels to represent objects, hand sequences, and object depth maps.

\subsubsection{Implementation Details}
The pre-trained weight of MimicMotion~\cite{zhang2024mimicmotion} is used for basic pose-driven human animation. We freeze VAE, CLIP, and DINO, training all other parameters on seven gpus with 40 GB. In the first stage, we train for 12,000 iterations at a resolution of ${512\times768}$ with 10 frames per sample. In the second stage, we train for 4,000 iterations at ${576\times1024}$ with six frames per sample. During these two processes, 65\% of the training data is sampled from our recorded dataset, while 35\% is sampled from open-source datasets. 
During the fine-tuning phase, we train five objects simultaneously. To enhance the generalization capability for face generation and mitigate the impact of limited human diversity, 80\% of the training data is sampled from the fine-tuning dataset, while the remaining 20\% comes from DreamDance.
The model undergoes 3,500 iterations at a resolution of ${576\times1024}$, requiring approximately three to four hours. The learning rate is set to \(10^{-5}\) with a 300-iteration warm-up. 
The HOI-region reweighting loss is applied with $\eta=1$ to enhance object detail learning.

During inference, CFG is set to 4.0. And we apply the time-aware position shift fusion from Sonic~\cite{ji2025sonic} to generate long videos, processing ten frames per sample. Inference for a 100-frame video takes five minutes and requires 22.84 GiB of VRAM.

\subsubsection{Comparison Methods}
We compare AnchorCrafter with five SOTAs: MimicMotion~\cite{zhang2024mimicmotion}, StableAnimator~\cite{chai2023stablevideo}, UniAnimate-DIT~\cite{wang2025unianimate}, Vace~\cite{vace}, and FlexiAct~\cite{zhang2025flexiact}. MimicMotion and StableAnimator are pose-guided human video generation methods based on UNet. UniAnimate-DIT is a digital human synthesis model leveraging the DiT architecture, while Vace is an all-in-one DiT-based model designed for both video creation and editing. FlexiAct is an advanced motion transfer framework that requires fine-tuning for specific actions. Consequently, we conducted a qualitative analysis to compare its performance. As these models do not directly accept object inputs, we prepared human images with objects pasted onto them for comparisons.
\subsubsection{Metrics}
We evaluate our system across multiple perspectives, including video quality, object shape, and appearance understanding, ensuring a comprehensive assessment of interaction realism and detail preservation. Using VBench~\cite{huang2024vbench} to measure subject consistency (Subj-Cons) and background consistency (Back-Cons) as well as motion smoothness (Mot-Smth). For object generation, we introduce Object-IoU, which utilizes SAM2 to extract object motion trajectory masks and compute the intersection over union (IoU) with the ground-truth mask. Additionally, Object-CLIP evaluates appearance consistency by computing the CLIP cosine similarity between generated objects and reference objects. For human generation, we use AdaFace~\cite{kim2022adaface} to extract facial features and calculate the cosine similarity between the generated face and the reference face, denoted as face-cos. We measure hand consistency using Landmark Mean Distances (LMD)~\cite{huang2024makeyouranchor}, leveraging OpenPose~\cite{cao2017openpose}.

\subsection{Main Results}
\label{sec:Main Results}
\begin{figure*}
  \centering
    \includegraphics[width=1.\linewidth]{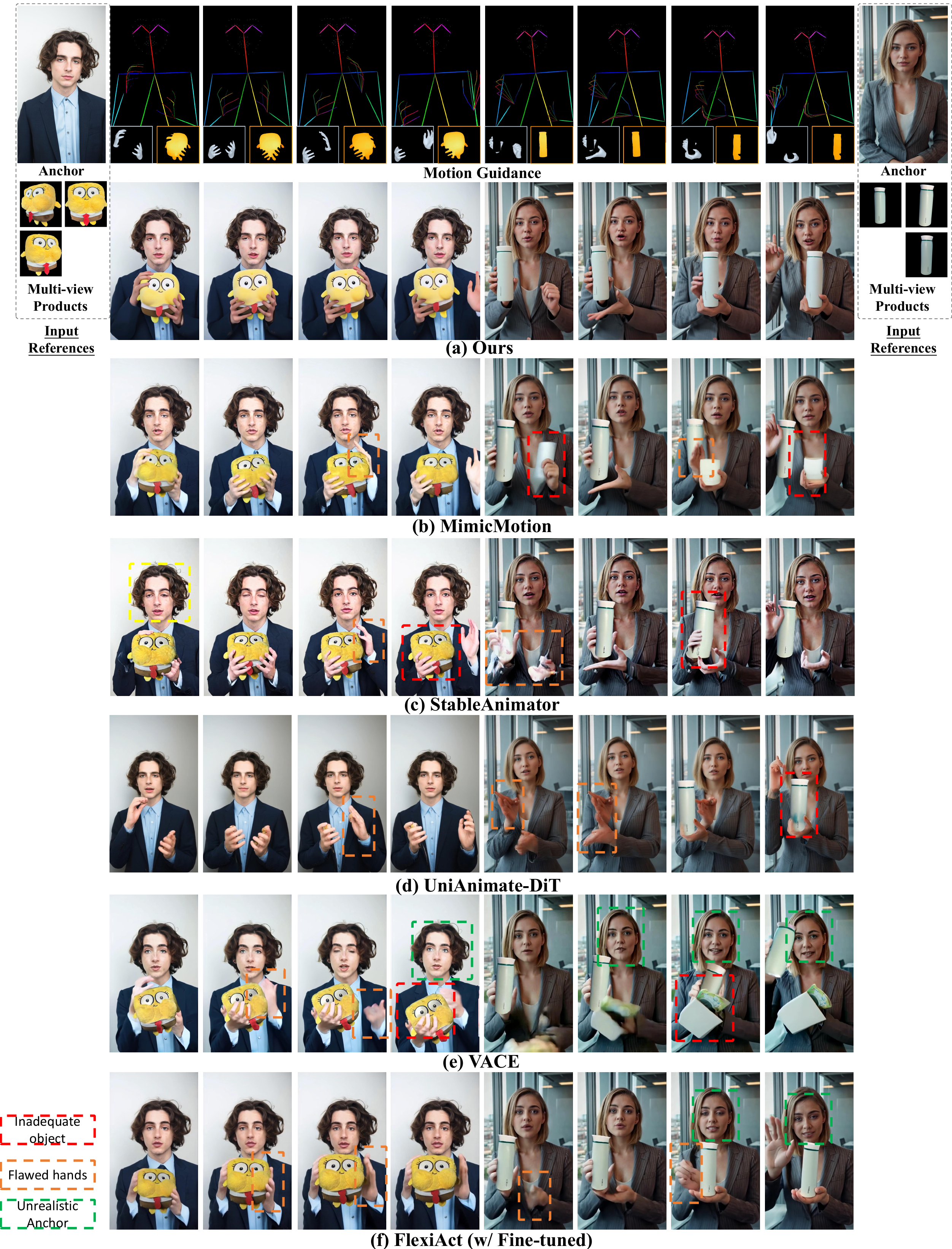}
   \caption{
Qualitative comparisons with other methods. Other approaches fail to preserve the visual integrity of individuals and objects or maintain reasonable interactive motion, making them incapable of completing character interaction tasks.
   }
  \label{fig:main_exper}
\end{figure*}




\begin{table*}[!t]
  \centering
    \caption{The quantitative results of our method were compared with those of SOTAs and ablation studies. Our method significantly outperforms existing approaches regarding numerical performance for spatial movement and appearance preservation of objects while also matching or exceeding current methods in image and video quality and human pose control capability. Subject consistency (Subj-Cons) and background consistency (Back-Cons) are percentages.}

  \resizebox{1\linewidth}{!}{
  \begin{tabular}{@{}lcc|cc|ccc@{}}
    \toprule 
    \multirow{2}{*}{Method}   & \multicolumn{2}{c}{Object}   & \multicolumn{2}{c}{Human}           & \multicolumn{3}{c}{Overall}\\

           & Obj-IoU↑     & Obj-CLIP↑    &Face-Cos↑      & LMD (Hand)↓  &Subj-Cons↑    &Back-Cons↑    &Mot-Smth↑\\
\midrule
MimicMotion    & 0.324        & 0.843        &0.60        & 20.43        & 93.24        &93.23         &98.62 \\
StableAnimator & 0.341        & 0.846        &0.55         & 19.14        & 92.95        &91.96         &98.38\\
UniAnimate-DiT &  -           &  0.772       &\textbf{0.73}& -          &  93.73          &  91.73            &98.88 \\
VACE           & 0.411        & 0.857        &0.40         & 22.78        & 91.50        &91.44         &98.55\\
Ours          &\textbf{0.906}&\textbf{0.921}&0.70&\textbf{12.19}&\textbf{95.43}&\textbf{94.96}&\textbf{98.91}\\

    \midrule
    w/o Human-Object Dual Adapter     &0.908	&0.892	&0.62	&12.33	&95.30	&94.75	&98.91    \\
    w/o Multi-View Self-Attention   &0.894	&0.904	&0.69	&12.41	&95.37	&94.74	&98.87\\
    w/o HOI-Motion Injection     &0.903	&0.918	&0.68	&13.19	&95.88	&95.13	&98.92  \\
    w/o HOI-Region Reweighting Loss   &0.895	&0.904	&0.65	&12.46&	95.40	&94.91	&98.94 \\   
    w/o Fine-Tuning             & 0.912	&0.847	&0.71	&12.83	& 94.75	&94.32	&98.85\\  
    \bottomrule
  \end{tabular}
  }
  \label{tab:exper}
\end{table*}

\subsubsection{Quantitative Results}
Table~\ref{tab:exper} presents the quantitative evaluation of our approach. Due to the complexity of UniAnimate-DiT’s alignment strategy, ground truth for pose is unavailable. Consequently, we did not assess the accuracy of hand and object positioning. 
The results highlight the superior performance of our method in object appearance reproduction, object motion, and hand generation. 

In terms of object motion, we achieves a significantly higher Object-IoU compared to competing methods, while other models fail to generate plausible object trajectories. Regarding object appearance preservation, we attains the highest Object-CLIP score among all evaluated approaches. Notably, some models attained relatively high scores; however, this was largely due to their tendency to treat objects as extensions of human clothing or background, causing them to remain static rather than as distinct, independent entities.

For human body generation, our method excels in both facial and hand synthesis. Our Face-Cos metric significantly surpasses that of the baseline model MimicMotion and is comparable to UniAnimate-DiT. This indicates that introducing object concepts did not compromise human generation quality.

Furthermore, VBench evaluations affirm that our generated videos exhibit superior subject consistency, background stability, and motion smoothness. These findings collectively validate the high visual quality and coherence of our approach.
\subsubsection{Qualitative Results}
Our qualitative results, illustrated in Fig.~\ref{fig:main_exper}, demonstrate the efficacy of our approach. In these figures, red boxes denote insufficient object appearance fidelity, orange boxes highlight inaccuracies in hand synthesis, and green boxes indicate unrealistic anchor placements. The examples on the left pertain to stationary object interactions, whereas those on the right involve object displacement. Our method achieves superior preservation of both human and object appearances while faithfully capturing their dynamic behaviors.

MimicMotion and StableAnimator generate high-quality, pose-driven human images; however, they exhibit coupling artifacts, treating objects as static extensions of clothing rather than independent entities. When displacement occurs, objects either remain fixed in place or exhibit visual artifacts. UniAnimate-DiT maintains human appearance relatively well but suffers from object disappearance in certain frames and produces the least accurate hand synthesis. VACE struggles to preserve human appearance, leading to severe object deformation. Due to the complexity of hand-object interactions, all models except ours fail in generating realistic hand representations. For FlexiAct, we conducted 4,000 rounds of fine-tuning on the corresponding test videos, enabling it to fully learn object motion and appearance. While it successfully achieves object displacement, the overall visual quality remains suboptimal, and it fails to transfer interactive motions to novel subjects.

By leveraging HOI-appearance perception and HOI-motion injection, our method effectively preserves both the appearance and motion dynamics of humans and objects, producing high-fidelity human-object interaction videos.
\subsection{Ablation Study}
\label{sec:Ablation Study}

\begin{figure}[htp]
\centering
    \centering
            \centering
            \includegraphics[width=\linewidth]{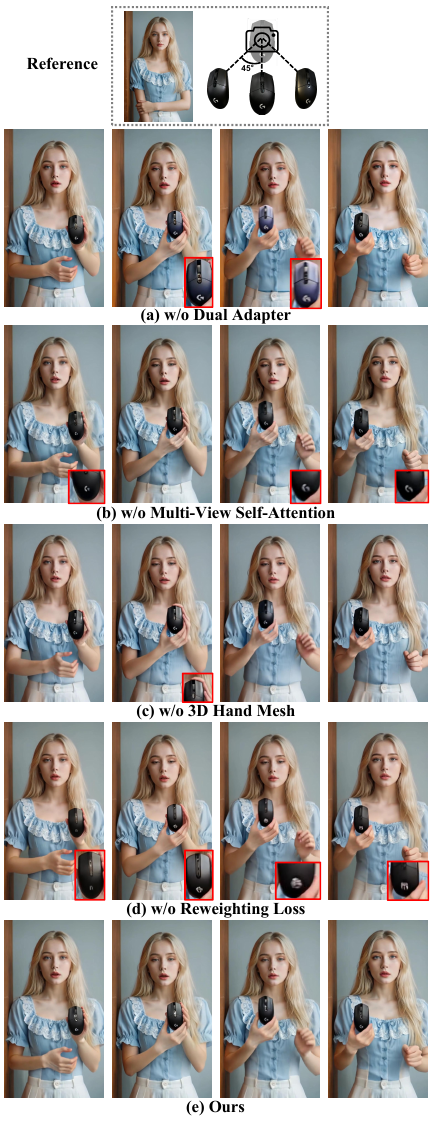}

\caption{
Ablation studies. Our modules improve the preservation of the object and its interactions with the hands. The error section has been enlarged in the lower right corner.
}
\label{fig:ablation}
\end{figure}

\noindent\textbf{Validation of human-object dual adapter.} 
We conducted detailed ablation experiments on human-object dual adapter. As shown in Table~\ref{tab:exper}. The Object-IoU metric remains superior to other models, which can be attributed to the detailed object trajectory guidance. However, the Object-CLIP score decreased, suggesting that merely injecting the latent representation of objects is insufficient for object perception, and the coupling with human subjects leads to a decline in face-cos. Fig.~\ref{fig:ablation} presents qualitative results, demonstrating that the human and background significantly influence objects. The human-object dual adapter effectively learns object representations and decouples human and object features. 

\noindent\textbf{Validation of multi-view self-attention.}
As shown in Table~\ref{tab:exper}, the removal of the multi-view self-attention branch results in a slight decrease in object CLIP performance, as multi-view self-attention facilitates the extraction of more object information. Since the changes occur at a fine-grained level, they are difficult for CLIP to capture, leading to only a minimal drop in the score. Additionally, as illustrated in Fig.~\ref{fig:ablation}, when the object's perspective shifts significantly, its logo undergoes deformation. Multi-view self-attention facilitates the extraction of more object information, effectively adapting to different viewpoints and accurately reconstructing the object's appearance.

\noindent\textbf{Validation of HOI-motion injection.}
We removed the hand 3D mesh information injection from the HOI-motion injection. As shown in Table~\ref{tab:exper}, the absence of detailed hand guidance resulted in a decrease in LMD (hand). Since errors are typically localized to specific fingers and are difficult for OpenPose to fully detect, the overall decline is not significant. The qualitative results illustrated in Fig.~\ref{fig:ablation} show that the complexity of hand-object interactions, coupled with inaccurate hand pose guidance provided by DWPose, leads to the generation of artifacts. These artifacts become particularly noticeable when the hand is occluded by objects.

\noindent\textbf{Validation of HOI-region reweighting loss.}
In conjunction with the HOI-region reweighting loss, we assigned higher weights to the hand-object regions during training to enhance the learning of objects and hands. As shown in Table~\ref{tab:exper}, insufficient emphasis on the hand-object regions causes the model to overfit to the human figures in the training set while failing to adequately learn object features, leading to inaccuracies in the appearance of both the object and the human figure. Fig.~\ref{fig:ablation} further demonstrates that, without adequate region enhancement, the model struggles to capture fine object details effectively.

\begin{figure}[htp]

\centering
    \centering
            \centering
            \includegraphics[width=.7\linewidth]{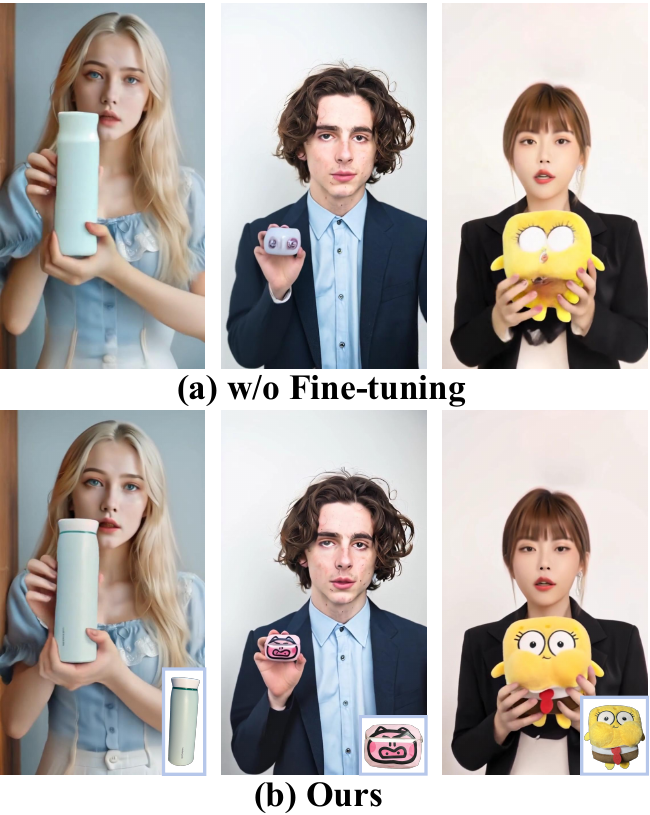}

\caption{
Ablation study of fine-tuning. After fine-tuning, the model can learn the texture details of objects.
}
\label{fig:abla-notune}
\end{figure}


\noindent\textbf{Validation of Fine-Tuning}
During the fine-tuning phase, we focus on learning the intricate texture details of objects. Table~\ref{tab:exper} demonstrates that fine-tuning significantly enhances the Object-CLIP score. As illustrated in Fig.~\ref{fig:abla-notune}, when presented with a new object, the model successfully generates appropriate interactive actions, along with its contour and primary color. However, it lacks detailed texture representation. This observation indicates that our foundational model already possesses the capability for human-object interaction. Through fine-tuning, the model further develops the ability to perceive and represent fine-grained texture details.

\subsection{User Study}
\label{sec:User Study}
\begin{table}[!t]
  \centering
  \caption{User study scores. The rating score is on a scale from one to five, where five is the highest score and one is the lowest.}
  \resizebox{\linewidth}{!}{
  \begin{tabular}{@{}lcc|cc|c@{}}
    \toprule
    \multirow{2}{*}{Method}                   & \multicolumn{2}{c}{Appearance} & \multicolumn{2}{c}{Motion}     &\multirow{2}{*}{Overall}    \\
                                              & Human      & Object            & Human      & Object                          \\
    \midrule
    MimicMotion~\cite{zhang2024mimicmotion}     &3.52         &1.84            & 3.68  & 2.08                     &3.08              \\
    StableAnimator ~\cite{tu2025stableanimator} & 3.02        &2.18           & 3.64    & 1.96                    &2.96   \\
    UniAnimate-DiT ~\cite{wang2025unianimate}  & 4.02         & 1.28          & 3.34     & 1.56                   &3.14   \\
    VACE ~\cite{vace}                           & 2.28         &1.92          & 2.42     & 2.16                   &2.62   \\
    Ours                                      &\textbf{4.12}& \textbf{4.68}     & \textbf{4.58} & \textbf{4.82}   & \textbf{4.64}     \\
    
    \bottomrule
  \end{tabular}}

  \label{tab:userstudy}
\end{table}

We conducted a user preference evaluation comparing AnchorCrafter with four state-of-the-art methods. Each participant reviewed 30 randomly selected videos for each method, scoring them based on five criteria:
\begin{itemize}
\item \textbf{Appearance Preservation (Human)}: Evaluates the consistency between humans and the reference images. 
\item \textbf{Appearance Preservation (Object)}: Focuses on maintaining object details such as texture and shape.
\item \textbf{Motion Accuracy (Human)}: Assesses the precision of human movements, particularly poses and gestures.
\item \textbf{Motion Accuracy (Object)}: Ensures that objects' motions align with intended physics and interactions.
\item \textbf{Overall Quality}: Provides a comprehensive evaluation of visual appeal, coherence, and stability.
\end{itemize}
Each criterion was rated on a five-point scale, with five being the highest score. Before the evaluation, participants were given detailed instructions, reference videos, and corresponding conditions to ensure fairness and clarity throughout the process. A total of 50 participants were invited, including 25 males and 25 females, achieving balanced gender representation. The participants consisted of 30 computer vision researchers, 10 artists, and 10 individuals from diverse backgrounds. This diverse group provided comprehensive feedback from various perspectives. As shown in Table~\ref{tab:userstudy}, our approach consistently ranked the highest across all five criteria, excelling particularly in motion accuracy and appearance diversity. This demonstrates its strength in generating realistic and stable human-object interaction videos.

\begin{figure}[htp]

\centering
    \centering
            \centering
            \includegraphics[width=1\linewidth]{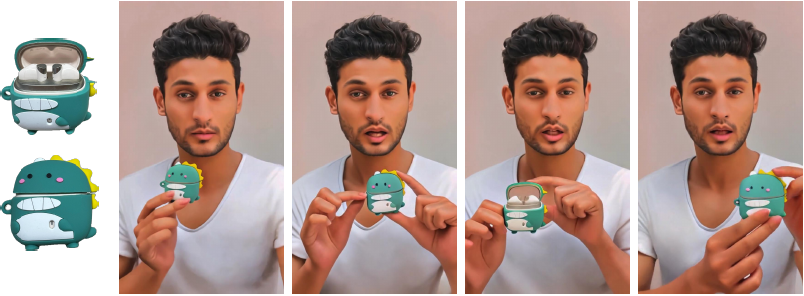}

\caption{
Anchorcrafter is able to perform the action of "opening the headphone case", demonstrating the adaptability and interactive ability of our model in complex scenarios.
}
\label{fig:fei}
\end{figure}

\subsection{Discussions}
\label{sec:Discussions}
\subsubsection{Generalization in Multi-View Tasks}
Human-object interaction tasks frequently require displaying objects from diverse perspectives. To this end, the proposed method utilizes multiple images to capture intricate details from various viewpoints. For simplicity, the experiment emphasizes the front view and two side views of the object. In practice, these views can be substituted with other perspectives during fine-tuning, such as the object's rear or even its interior. As shown in Fig.~\ref{fig:fei}, leveraging depth perception and multi-view images, we successfully achieve the complex task of "opening the headphone case". This approach effectively captures detailed features from various angles of the headphones, showcasing our model's adaptability and interaction capabilities in intricate scenarios.

\subsubsection{Generalization Across Shape-Similar Objects}

\begin{figure}[!t]
\newcommand{\galleryfigurewidth}{0.26}
\centering

    \centering

            \centering
            \includegraphics[width=.8\linewidth]{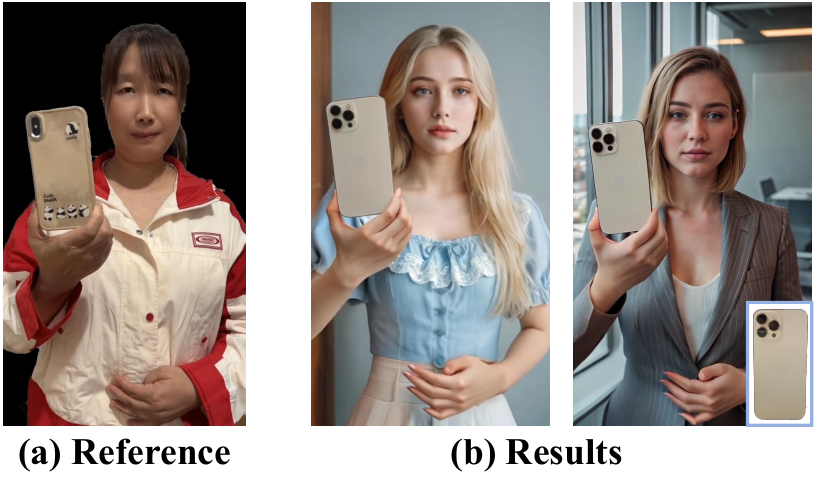}

\caption{
Generalization over similar object shapes. Driving iPhone interaction videos with another mobile phone.}
\label{fig:similar}
\end{figure}
During inference, generate interaction videos with a specific object but don't have access to it, can use a similarly shaped object instead. By capturing footage with a similar object, we can extract human poses, 3D hand mesh, and object depth as conditioning inputs. Our model exhibits a certain degree of generalization capability, enabling accurate inference even with similar-shaped objects. As shown in Fig.~\ref{fig:similar}, we utilized videos recorded with different phone models to drive video generation for an iPhone. For objects with similar shapes, template videos can be captured to guide multiple objects.
\begin{figure}[!t]
  \centering

   \includegraphics[width=0.9\linewidth]{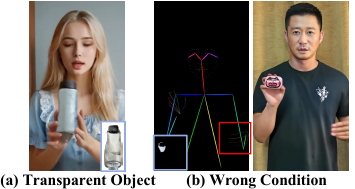}
   \caption{
   Limitations. Handling transparent objects remains challenging, and erroneous conditions introduce interference.
   }
  \label{fig:limit}
\end{figure}
\subsection{Limitations} AnchorCrafter is primarily evaluated on rigid objects and exhibits suboptimal performance when handling transparent objects. Additionally, inaccuracies in preprocessing models such as DWPose introduce errors into the generated results. As shown in Fig.~\ref{fig:limit} (left), the model fails in cases of mirror penetration. In Fig.~\ref{fig:limit} (right), the 3D hand mesh provides accurate hand guidance, whereas DWPose incorrectly assigns the left hand, leading to erroneous generation results. In future work, we will conduct a more in-depth exploration of the challenges associated with transparent and non-rigid objects.

\section{Conclusion}
\label{sec:conclusion}
We present AnchorCrafter, a novel diffusion-based system for anchor-style product promotion video generation that incorporates human-object interaction into pose-guided human video generation. Our system addresses key challenges in object motion guidance, appearance preservation, and complex human-object interactions by introducing HOI-appearance perception and HOI-motion injection. We also propose an HOI-region reweighting loss to improve object details during training. Extensive experiments show that AnchorCrafter outperforms existing methods, achieving superior object appearance preservation and shape awareness while ensuring high-quality video generation with consistent human appearance and motion.

{
    \small
    \bibliographystyle{IEEEtran}
    \bibliography{main}
}

\end{document}